\documentclass[10pt,twocolumn,letterpaper]{article}

\usepackage{wacv}
\usepackage{times}
\usepackage{epsfig}
\usepackage{graphicx}
\usepackage{amsmath}
\usepackage{amssymb}
\usepackage{booktabs}
\usepackage{tabularx}
\usepackage{makecell}
\usepackage{lineno}
\usepackage{multirow}

\usepackage[symbol]{footmisc}
\def\wacvPaperID{2123} 

\wacvfinalcopy 

\ifwacvfinal
\usepackage[breaklinks=true,bookmarks=false]{hyperref}
\else
\usepackage[pagebackref=true,breaklinks=true,colorlinks,bookmarks=false]{hyperref}
\fi

\usepackage{amsmath}
\usepackage{amssymb}
\usepackage{mathtools}
\usepackage{amsthm}
\usepackage{pifont}
\usepackage[capitalize,noabbrev]{cleveref}
\usepackage{mathtools}
\usepackage{algorithm}
\usepackage{algpseudocode}
\usepackage{float}
\usepackage{color}
\usepackage{colortbl}
\usepackage{rotating}
\usepackage{times}
\usepackage{epsfig}
\usepackage{enumitem}
\usepackage{makecell}
\usepackage{lineno}
\usepackage{tabularx}
\usepackage{xcolor}
\usepackage{multirow}
\usepackage{graphicx}
\usepackage{hhline}
\usepackage{textcomp,booktabs}
\usepackage{caption}
\usepackage{stmaryrd}
\usepackage[mathscr]{eucal}
\usepackage{lineno}
\usepackage[export]{adjustbox}

\def\0{{\bf 0}}
\def\1{{\bf 1}}

\theoremstyle{plain}

\theoremstyle{definition}

\theoremstyle{remark}

\usepackage[textsize=tiny]{todonotes}

\definecolor{red}{rgb}{0.95,0.4,0.4}
\definecolor{purered}{rgb}{1,0,0}
\definecolor{blue}{rgb}{0.4,0.4,0.95}
\definecolor{darkblue}{rgb}{0,0,0.8}
\definecolor{darkred}{rgb}{1,0,0}
\definecolor{darkgreen}{rgb}{0,0.5,0}
\definecolor{grey}{rgb}{0.6,0.6,0.6}
\definecolor{col1}{RGB}{232, 161, 148}
\definecolor{col2}{RGB}{148, 187, 232}
\definecolor{lightgrey}{rgb}{0.85,0.85,0.85}
\definecolor{lightlightgrey}{rgb}{0.9,0.9,0.9}
\definecolor{verylightBG}{rgb}{0.9,0.99,0.99}
\definecolor{darkgreen}{rgb}{0.3, 0.75, 0.3}

\newcommand{\cmark}{\ding{51}}%

\begin{document}

\title{Far3Det: Towards Far-Field 3D Detection}

\author{
Shubham Gupta\textsuperscript{1$\ast$} \quad
Jeet Kanjani\textsuperscript{1}\thanks{Co-first authors.  $^\dag$Co-last authors.} 
\quad
Mengtian Li\textsuperscript{1} \quad
Francesco Ferroni\textsuperscript{2} \\
James Hays\textsuperscript{2,3} \quad
Deva Ramanan\textsuperscript{1,2 $\dag$}
\quad\quad
Shu Kong\textsuperscript{4 $\dag$}
\\ 
\textsuperscript{1}CMU \quad\quad  
\textsuperscript{2}Argo AI  \quad\quad  
\textsuperscript{3}Gatech  \quad\quad  
\textsuperscript{4}Texas A\&M University 
\vspace{-3mm} 
}

\maketitle
\thispagestyle{empty}

\begin{abstract}
We focus on the task of far-field 3D detection (Far3Det) of objects beyond a certain distance from an observer, e.g., $>$50m. Far3Det is particularly important for autonomous vehicles (AVs) operating at highway speeds, which require detections of far-field obstacles to ensure sufficient braking distances. However, contemporary AV benchmarks such as nuScenes underemphasize this problem because they evaluate performance only up to a certain distance (50m). One reason is that obtaining far-field 3D annotations is difficult, particularly for lidar sensors that produce very few point returns for far-away objects. Indeed, we find that almost 50\% of far-field objects (beyond 50m) contain zero lidar points. Secondly, current metrics for 3D detection employ a ``one-size-fits-all" philosophy, using the same tolerance thresholds for near and far objects, inconsistent with tolerances for both human vision and stereo disparities. Both factors lead to an incomplete analysis of the Far3Det task. For example, while conventional wisdom tells us that high-resolution RGB sensors should be vital for 3D detection of far-away objects, lidar-based methods still rank higher compared to RGB counterparts on the current benchmark leaderboards. As a first step towards a Far3Det benchmark, we develop a method to find well-annotated scenes from the nuScenes dataset and derive a well-annotated far-field validation set. We also propose a Far3Det evaluation protocol and explore various 3D detection methods for Far3Det. Our result convincingly justifies the long-held conventional wisdom that high-resolution RGB improves 3D detection in the far-field.
We further propose a simple yet effective method that fuses detections from RGB and lidar detectors based on non-maximum suppression, which remarkably outperforms state-of-the-art 3D detectors in the far-field.
\end{abstract}

\section{Introduction}
\label{sec:intro}

\begin{figure*}
\centering
\includegraphics[width=1.0\textwidth]{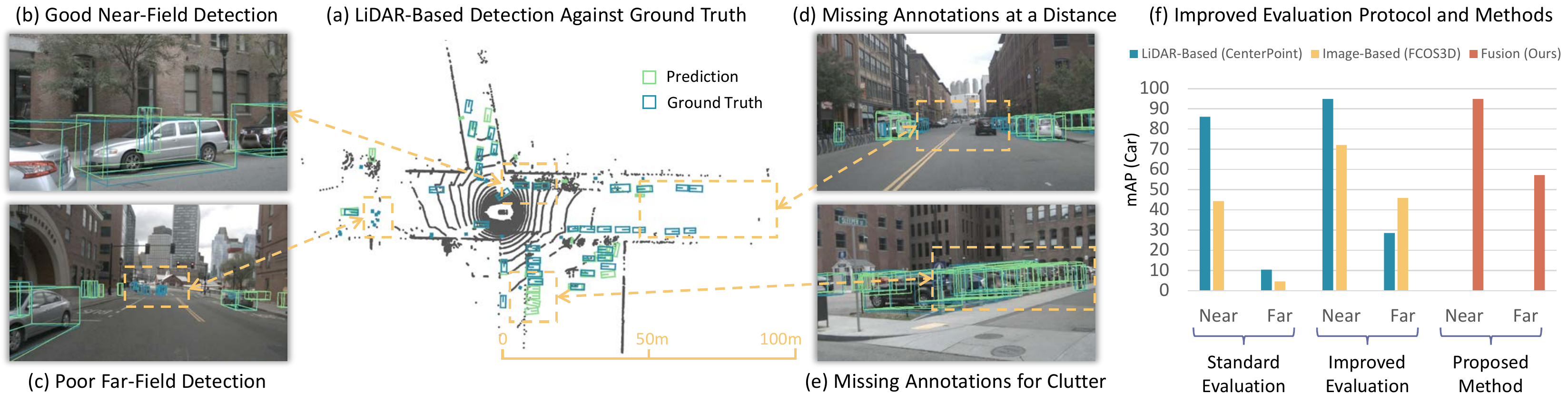}
\vspace{-7mm}
\caption{\small
We study the problem of far-field 3D detection (Far3Det)  of objects beyond a certain distance from an observer ({\bf a}). Far3Det is crucial to self-driving safety because autonomous vehicles (AVs) must detect far-field objects to avoid a potential collision. The field of 3D detection has been greatly advanced by modern benchmarks, where lidar-based detectors prove to be superior to image-based detectors.
However, these benchmarks evaluate detections up to a certain distance (50m), masking the poor far-field object detection performance of lidar-based detectors ({\bf b} \& {\bf c}). Even worse, existing benchmarks do not sufficiently annotate far-field objects ({\bf d} \& {\bf e}), partly due to fewer lidar returns.
To study Far3Det, we derive a reliable validation set and set up benchmarking protocols. We analyze well-established methods ({\bf f}) and justify the conventional wisdom (for the first time, quantitatively) that using high-resolution RGB boosts Far3Det.
We explore multimodal detectors by fusing RGB- and lidar-based detections. We propose a rather simple fusion method based on non-maximum suppression adaptive to distance,  achieving significant improvement over the state-of-the-art lidar-based detectors.
}
\vspace{-1mm}
\label{fig:teaser}
\end{figure*}

Autonomous vehicles (AVs) must detect objects \emph{in advance} for timely action to ensure driving safety~\cite{vanderbilt2009traffic,king2017top}. Because a 60 mph vehicle requires 60 meter stopping distance~\cite{greibe2007braking}, AVs must detect far-field obstacles to avoid potential collisions. Additionally, detecting far-field objects is also relevant for navigation in urban settings at modest speeds during precarious maneuvers, such as unprotected left turns where opposing traffic might be moving at 35mph, resulting in a relative speed of 70mph~\cite{joshua1992mitigation}. 
These real-world scenarios motivate us to study the problem of far-field 3D object detection ({\bf Far3Det}). Fig.~\ref{fig:teaser} previews this work.

\textbf{Status quo.}
3D detection has greatly advanced under AV research, largely owing to contemporary industry-level benchmarks  that collect data using lidar (e.g., nuScenes \cite{Caesar_2020_CVPR}, Waymo~\cite{sun2020scalability}, and KITTI~\cite{Geiger2012CVPR}), which helps precise annotation / localization in the 3D world. 
These benchmarks evaluate detections only up to a certain distance (i.e., within 50 meters from the ego vehicle)~\cite{Geiger2012CVPR,Caesar_2020_CVPR,sun2020scalability}, because of the difficulty in annotating far-field objects with few or zero lidar returns.
This limits the exploration of Far3Det methods, and even leaves unjustified the conventional wisdom that RGB processing should improve detection in the far field~\cite{xu2018pointfusion,meyer2019sensor,liang2018deep,xu2021fusionpainting}.
We demonstrate the reasons for this disparity include both lack of high-quality far-field annotations and the lack of range-aware metrics.

\textbf{Why is Far3Det hard?}
Precisely localizing far-field objects in the 3D world is difficult even for humans~\cite{welchman2004human,rushton2009observers}. 
Perceptually, human drivers are able to detect far-field objects but may not be able to report their precise 3D locations. We argue that evaluation of Far3Det needs new range-based metrics to serve autonomous driving.
Moreover, in terms of 3D sensor technology,
while lidar has proven incredibly effective for near-to-mid range, it produces notoriously sparse outputs for long range perception~\cite{Caesar_2020_CVPR}; indeed, it may not even return points for distant objects. Indeed,
we find that almost 50\% of far-field objects (beyond 50m) in conventional benchmarks contain zero
lidar points. 
As past work has shown, different sensors (such as RGB cameras) can produce higher-resolution data which is more effective for far-field perception, suggesting that multimodal processing~\cite{xu2018pointfusion,qi2018frustum} will be crucial for Far3Det.

{\bf Annotation and evaluation.}
One reason for the limited exploration of Far3Det is the difficulty in annotating far-field objects (since lidar has particularly sparse returns on far-field objects).
As a result, existing benchmarks lack sufficient far-field object annotations~\cite{Caesar_2020_CVPR}.
To obtain a high-quality annotated validation set for fair Far3Det benchmarking, we analyze the well-established nuScenes dataset~\cite{Caesar_2020_CVPR} and develop a method to derive a subset that has high quality annotations in the far field (Fig.~\ref{fig:annotation_stats}). 
We call this set ``Far nuScenes'' and make it publicly available. We also analyze the set of far-away objects with zero-lidar points that are conventionally ignored. We first point out they comprise a sizeble fraction (15.6\% of all annotations!), and introduce an automatic procedure for identifying and {\em re}including the subset of them which are visible in at least one modality.
Finally, we notice that typical benchmark metrics such as 3D mAP use a fixed distance threshold (e.g., 0.5,1,2 and 4m)~\cite{Caesar_2020_CVPR} which is too harsh for the far-field, while potentially too relaxed for the near-field. Therefore, we introduce new metrics based on adaptive thresholds w.r.t distance from ego-vehicle. The rationale is to use small thresholds for near-field objects and relaxed thresholds for the far-field. Finally, a natural question is whether there exists other datasets (besides nuScenes) that facilities the exploration of Far3Det. We acknowledge that recent datasets such as Argoverse 2.0~\cite{wilson2021argoverse} and Waymo Open dataset~\cite{sun2020scalability} do include annotations for objects beyond 50m. 
We choose to explore Far3D on nuScenes because this dataset (1) has become the standard benchmark for 3D detection, (2) contains a larger resource of off-the-shelf models for 3D detection across different modalities, and (3) has carefully annotated objects which are beyond the effective lidar range or occluded. Moreover, we posit that RGB sensors will continue be higher resolution than lidar sensors (due to the manufacturing process), and so argue that multimodal processing will continue to be valuable for pushing the limits of perceptual range, even for future sensor configurations.

{\bf Multimodal detection.} 
We believe that leveraging multimodal signals can improve 3D detection~\cite{xu2018pointfusion,meyer2019sensor,liang2018deep,xu2021fusionpainting}. For example, conventional wisdom states that high-resolution RGB better captures objects in the far field where lidar returns rather sparse points.
Although multimodal detection has been well studied on numerous benchmarks~\cite{8968513,ku2018joint}, leaderboards often reveal only marginal improvements for multimodal detectors over their single-modal lidar-based counterparts. We posit, and verify by experiment, that with proper range-based localization metrics and high-quality far-field annotations, one can quantitatively justify (for the first time, to our knowledge) the conventional wisdom that multimodal detection is crucial for far-field detection. To demonstrate this, we introduce a simple but effective NMS method for fusing RGB-only detections with state-of-the-art lidar-only detections, considerably improving far-field accuracy. We also combine our fusion approach with recent work of~\cite{yin2021multimodal}, demonstrating significant improvements for far-field 3D object detection.

{\bf Contributions.} 
We make three major contributions for the study of far-field 3D detection (Far3Det).
Firstly, we propose a method to find well-annotated scenes in the well-established nuScenes dataset and derive a new validation set for fair Far3Det benchmarking, together with our new range-based metrics.
Secondly, we extensively study various detectors including recent state-of-the-art RGB-lidar fusion networks, justifying the conventional wisdom that using RGB boosts Far3Det.
Thirdly, we propose a rather simple yet effective RGB-lidar fusion method based on non-maximum suppression (NMS) that fuses detections w.r.t distances. Our method remarkably outperforms the state-of-the-art lidar detectors and serves as a baseline for future research of Far3Det.

\begin{figure*}[t]
\centering
\includegraphics[width=1\linewidth, clip, trim={0cm 0.4cm 0cm 0.2cm}]{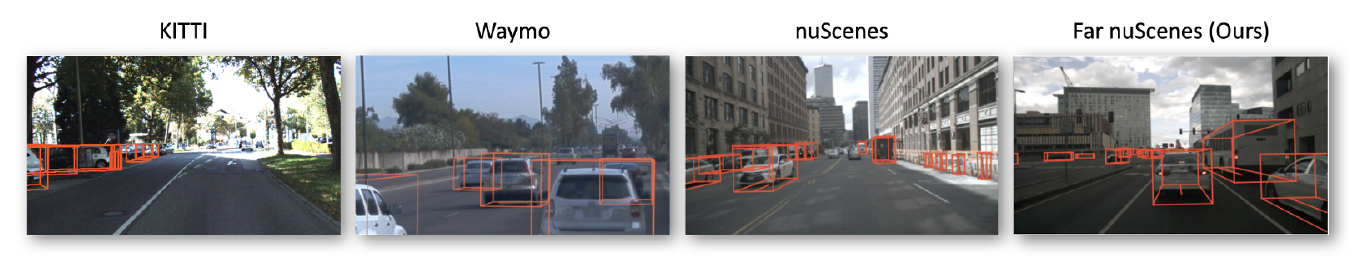}
\vspace{-7mm}
\caption{\small 
Ground-truth visualization for some well-established 3D detection datasets.
Clearly, they (the first three columns) have significant amount of missing annotations on far-field objects  (a quantitative analysis is in Fig~\ref{fig:annotation_stats}).
To obtain a reliable validation set, we describe an efficient verification process (in Sec~\ref{ssec:dataset_eval}) for identifying annotators that consistently produce high-quality far-field annotations. This helps us identify high-quality far-field annotations, and derive Far nuScenes (rightmost) that supports far-field detection analysis. 
}
\vspace{-4mm}
\label{fig:miss-annotation}
\end{figure*}

\section{Related Work}
\label{sec:related work}

{\bf 3D Detection Benchmarks.} 
There exist many excellent multimodal benchmarks for 3D detection in the context of autonomous driving, such as KITTI~\cite{Geiger2012CVPR}, Waymo Open Dataset~\cite{sun2020scalability}, and nuScenes~\cite{Caesar_2020_CVPR}. KITTI~\cite{Geiger2012CVPR} was the
pioneering multimodal dataset providing dense point cloud from a lidar sensor along with front-facing stereo images. Recently, Waymo Open dataset~\cite{sun2020scalability} and nuScenes~\cite{Caesar_2020_CVPR} were released which provide significantly more annotations than KITTI~\cite{Geiger2012CVPR} to further advance the research in the AV community.
All these benchmarks exclusively focus on near-field objects and ignore far-field ones.
A primary reason is that the far-field objects are hard to annotate.
As a result, they benchmark on detecting objects up to a certain distance from the observer (e.g., 50m).
Within this distance, objects have enough lidar returns, and lidar-based d, they outperform image-based detectors for the task of 3D detection.
Motivated by the safety concerns in AV research, we look far enough ``out'' by evaluating detection performance on far-field objects ($\ge$50m).

{\bf 3D Object Detection} 
aims to predict 3D bounding boxes (cuboids). In the AV field, the input to a 3D detector can be either lidar point cloud or an RGB image. Over a single 2D image, detecting objects and estimating their 3D characteristics is an exciting  topic~\cite{wang2019pseudo, wang2021fcos3d, peri2022towards}. 
This is known as Monocular 3D Object Detection~\cite{wang2021fcos3d} or image-based 3D detection~\cite{wang2019pseudo}.
We use the latter in this paper to contrast the lidar-based 3D detector.
In AV research, lidar-based 3D detectors prove to be a great success in terms of 3D understanding.
There are numerous 3D detectors built on lidar input~\cite{lang2019pointpillars,zhu2019class,wang2019exploit,hu2019wysiwyg,yan2018second,yin2020center}.
They greatly outperform image-based 3D detectors in the existing benchmarks~\cite{Caesar_2020_CVPR,sun2020scalability,Geiger2012CVPR},
presumably because lidar points are strong signals for precise 3D localization of near-field objects.
The Waymo~\cite{sun2020scalability} and Argoverse 2.0~\cite{wilson2021argoverse} datasts evaluate on far-field objects, but they do not explicitly compare the RGB-based models with the lidar-based models. Therefore, study of using RGB to improve Far3Det is more like a conventional wisdom without justification. Our work convincingly, quantitatively concludes that using high-resolution RGB boosts Far3Det.

{\bf 3D Detection over Multi-Modalities.}
Fusing multimodal data for 3D object detection is an active field. There are many approaches: some encode lidar and camera information separately and then fuse at an object-proposal stage~\cite{chen2017multiview, ku2018joint, peri2022towards};
some try to augment lidar points with either RGB features~\cite{sindagi2019mvxnet} or semantic information obtained by processing RGB inputs~\cite{vora2020pointpainting}; 
some work in reverse that augment RGB images with dense depth, guided by lidar measurements~\cite{you2020pseudolidar}; 
some work in a staged manner by first detecting boxes via image data and subsequently localizing in 3D with lidar~\cite{qi2018frustum}; 
and others focus on the late-stage fusion of detections from multimodal inputs~\cite{xu2018pointfusion}.
Probabilistic Fusion proposed by \cite{chen2021multimodal} is another simple non-learning based approach for late fusion of object detectors derived from first principles given conditional independence assumptions~\cite{henrion2013qualitative}. CLOCs is a recent learning-based multimodal detector for fusing detections computed from lidar and image modalities~\cite{pang2009clocs}. Recently, multi-view virtual points~\cite{yin2021multimodal} was introduced and achieved SOTA performance on nuScenes~\cite{Caesar_2020_CVPR} benchmark. 
The variety of multimodal detection methods indicates that this is still an active area of research and there is currently no single approach that significantly outperforms others.
Furthermore, many multimodal 3D detectors underperform the state-of-the-art (single-modal) lidar-based detectors in leaderboards~\cite{sun2020scalability,Caesar_2020_CVPR}.
Along with added complexity for fusion methods, we believe that this results in less focus towards fusion methods for 3D detection methods in AV. 
Our exploration of Far3Det demonstrates that RGB is quite useful and leveraging both RGB and lidar greatly improves detection performance especially at far-field.

\section{Far-Field 3D Detection}
\label{sec:methodology}

We now describe the far-field detection problem in detail. We explore various publicly available datasets for this problem. We set up the evaluation protocol for Far3Det where we introduce new, reasonable metrics.

\subsection{Dataset}
\label{ssec:dataset_eval}

As mentioned earlier, the primary reason why Far3Det was not explored (till now) is the difficulty in data annotation, i.e., it is hard to label 3D cuboids for far-field objects if they have few or no lidar returns.
Despite this difficulty, a reliable validation set is required for the study of Far3Det. One of our contributions is the proposed strategy to construct a well-annotated val-set derived from existing dataset.

\begin{figure}[t]
\centering
\includegraphics[width=0.48\linewidth, clip, trim={0cm 2.2cm 0cm 0cm}]{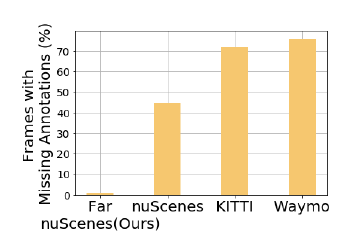}
\includegraphics[width=0.48\linewidth, clip, trim={0cm 0cm 0cm 2.4cm}]{imgs/anno-comp.pdf}
\vspace{-3mm}
\caption{\small (a) We randomly sample 50 frames from each dataset and manually inspect missing annotations for $>$50m objects to analyze the annotation quality of existing 3D detection datasets. This analysis suggests that the derived subset Far nuScenes has higher annotation quality compared to existing benchmarks, KITTI~\cite{Geiger2012CVPR}, Waymo~\cite{sun2020scalability} and nuScenes~\cite{Caesar_2020_CVPR}. 
(b) We compare the average number of annotations per frame at a given distance between Far nuScenes (yellow) over the standard nuScenes (blue), showing that the former (ours) has higher annotation density.
}
\label{fig:annotation_stats}
\vspace{-5mm}
\end{figure}


Today's 3D detection datasets (e.g., nuScenes~\cite{Caesar_2020_CVPR}, KITTI~\cite{Geiger2012CVPR} and Waymo~\cite{sun2020scalability}) emphasize  lidar because of its faithful geometric measure of 3D world. As lidar point density decreases with distance, they have poor annotations for objects in the far field, as shown in Fig.~\ref{fig:miss-annotation}, making the study of Far3Det almost impossible.
To foster the research in this space, we focus our efforts on cleaning up nuScenes, which has arguably reported higher far-field annotations than other publicly available datasets (see suppl).

{\bf Finding well-annotated far-field data.}
Manually inspecting individual annotations is prohibitively expensive. Based on past video annotation interfaces~\cite{vondrick2013efficiently,doermann2000tools}, we believe that individual annotators tend to be assigned to particular {\em scenes}, or sequential sweeps of 40 frames annotated at 0.5s intervals (every 2 frames). One insight from the crowdsourcing literature is that, while different annotators may be inconsistent, individual annotators are often self-consistent~\cite{nowak2010reliable}. We have verified that this assumption appears to hold in nuScenes; certain scenes tend to be consistently annotated with far-field objects compared to others.

Therefore, we design a pipeline for verifying annotators/scenes as follows. For each scene, we randomly sample 20 frames and mark any missing far-field annotations. We manually remove scenes for which more than 2 missing far-field annotations are found for any frame. We then collect all the remaining good scenes and ensure almost all far-field objects are annotated. We end up with a \emph{Far nuScenes} val-set that contains 38 scenes (out of 150 in the original nuScenes val-set). Our scene-based verification roughy reduces manual effort of data cleaning by 40$\times$ (compared to inspecting each lidar frame). 
Fig.~\ref{fig:annotation_stats} shows the statistics of Far nuScenes in comparison to other datasets. Also, to ensure that Far nuScenes is large-scale enough for benchmarking, we diligently sample diverse frames to cover scenes in the daytime and nights, in urban scenes and highways, etc.

\begin{figure}[t]
\centering
\includegraphics[width=0.8\linewidth, height=5.2cm, clip, trim={0cm 0cm 0cm 2.5cm}]{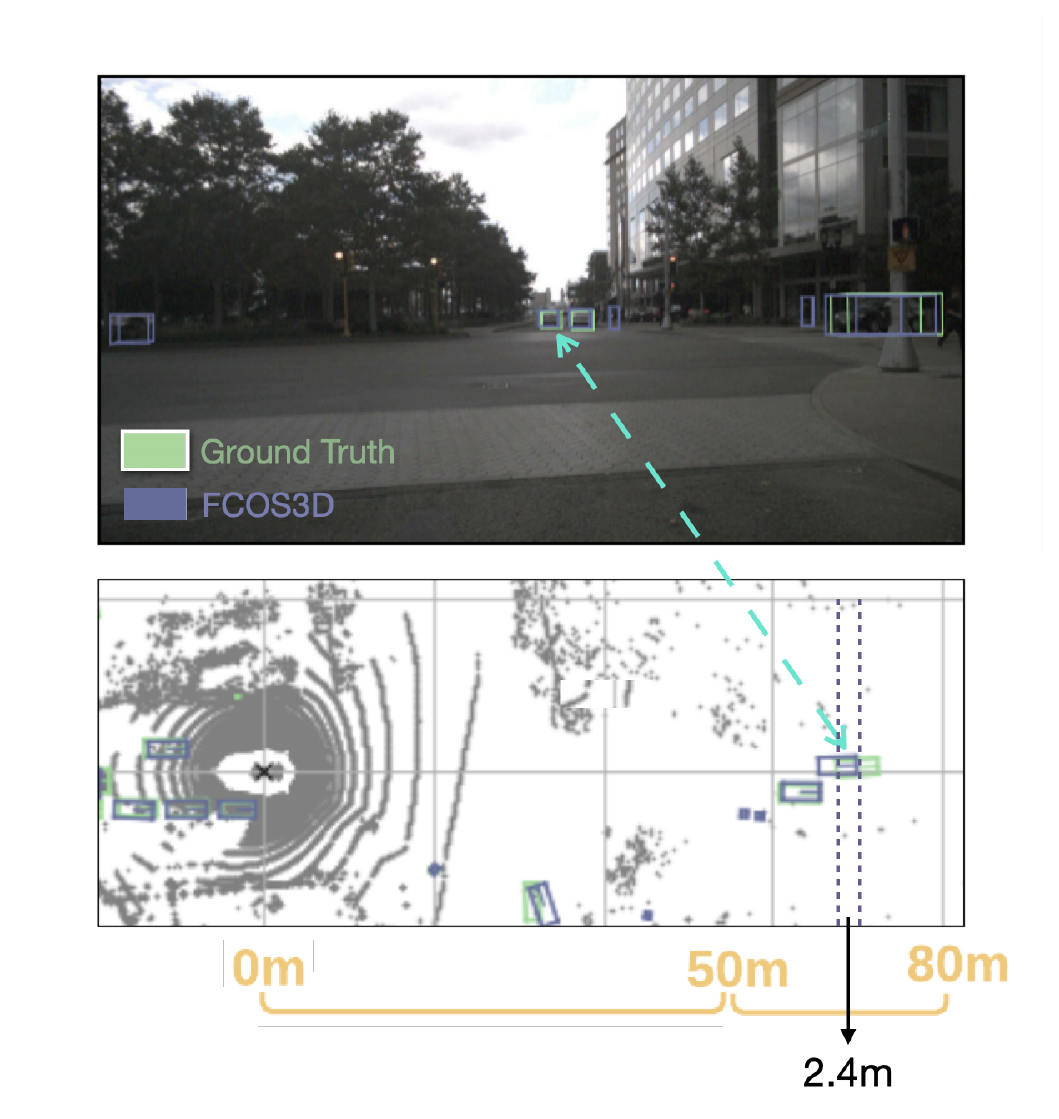}
\vspace{-6mm}
\caption{\small
Standard evaluation protocol evaluates detection using small distance thresholds (i.e., \{0.5, 1, 2, 4\} meters).
Image-based detector (FCOS3D) has a  typical ``failure'' mode as below.
The detections are quite good in the projected image space (top row) but might have notable errors in the BEV (bottom row). This evaluation protocol penalizes such errors too heavily.
}
\label{fig:metric_visual}
\vspace{-4mm}
\end{figure}


\subsection{Detection of Zero-lidar Point Objects}
\label{subsec: zero_points_long}
We found that $15.6\%$ annotated objects in nuScenes that have zero lidar points. 
Further analysis suggests that such situations  arise when (1) objects are occluded at the current frame but annotated based on previous or next frames, and (2) objects are too far that have no lidar returns (more details in appendix). 
One may ask how these objects were annotated. According to nuScenes documentation and its authors, such objects were annotated by interpolating annotations across frames (for occluded ones)~\cite{nuscenes:forum} and using RGB images as visual evidence. By default, nuScenes filters all the zero-lidar objects in evaluation. 
In this work, we \emph{re}include these zero-lidar objects for evaluation because detecting them is crucial for AV safety.

In order to identify these unoccluded objects, we define the following strategy. A box is tagged as occluded if (1) there is another box in front of it, or (2) there are raw lidar points (due to unannotated objects) in front of it. 
We start by projecting all the lidar points and annotated boxes to the image view for all the cameras. In our setting, we consider a box to be occluded if it has (1) more than 0.5 IoU overap with a box 5 meters or more in front, or (2) lidar points 10 meters or more in front. We tune these parameters by visualizing the projections in the image view. On performing this analysis, we find that the unoccluded zero lidar point objects account for one-third of the total zero-lidar Point objects which amounts to 5.44 \% of the total nuScenes dataset. We discard the occluded boxes and include the rest of the zero-lidar objects in our analysis (results in Sec.~\ref{subsec: zero_points}) as shown in Table~\ref{tab:zero-lidar-frac} .

{\setlength{\tabcolsep}{0.1em}
\begin{table}[t]
\scriptsize
\centering
\caption{\small 
Fraction of annotations with zero lidar points found in nuScenes Train and Val dataset for 0-80m. This metric is provided by considering the current lidar sweep as well as last 10 lidar sweeps. To calculate the latter, the annotations are interpolated between their previous and current position. 
}
\vspace{-2mm}
\adjustbox{max width=\linewidth}{
\begin{tabular}{c c c c}
\toprule
\multirow{2}{*}{\textbf{Data/0-80m}} & \multirow{2}{*}{Total Annotations} & Zero lidar point annotations & Zero lidar point annotations  \\
 &  & at the current frame & interpolated across ten frames\\
\cmidrule(r){1-1}  \cmidrule(r){2-2}  \cmidrule(r){3-3}  \cmidrule(r){4-4} 
\textbf{Train-set} &  944881 & 147702 (15.6\%) & 94990 (10.0\%)\\ 
\textbf{Val-set} & 187528 & 29275 (15.4\%) & 19007 (10.1\%)   \\ 
\hline
\end{tabular}
}
\vspace{-1mm}
\label{tab:zero-lidar-frac}
\end{table}
}



\subsection{Evaluation Protocol}
\label{sec:metric}

\textbf{Standard 3D Detection Metrics}.
Any typical 3D detection metric can evaluate Far3Det performance such as the mean average precision (mAP) over all classes. In each class, its AP averages the list of precisions varied by a threshold. In 3D detection, the center-distance based threshold is used to determine whether a detection matches a ground-truth object. The standard distance thresholds used in nuScenes~\cite{Caesar_2020_CVPR} are 0.5m, 1m, 2m, and 4m.

{\bf AP with Adaptive Distance Thresholds}. We demonstrate that for the Far3Det, it is seemingly harsh to penalize far-field localization errors using small distance thresholds (e.g., 0.5 meters).
In fact, we human beings find it extremely difficult (if not impossible) to localize far-field moving objects~\cite{welchman2004human,rushton2009observers}. As shown in Fig.~\ref{fig:metric_visual}, knowing an oncoming car's direction is more important than precisely locating it in the 3D world provided its distant from us.
For this reason, Waymo benchmark introduces a longitudinal error tolerant 3D average precision for RGB-only based detection~\cite{hung2022let}. In this paper, for general far 3D detection, we propose an adaptive thresholding scheme in which the threshold to match a detection with ground-truth annotations increases with distance. 


\begin{figure}[t]
\centering
\begin{minipage}[r]{0.22\textwidth}
\centering
\includegraphics[width=1\linewidth, height=4cm]{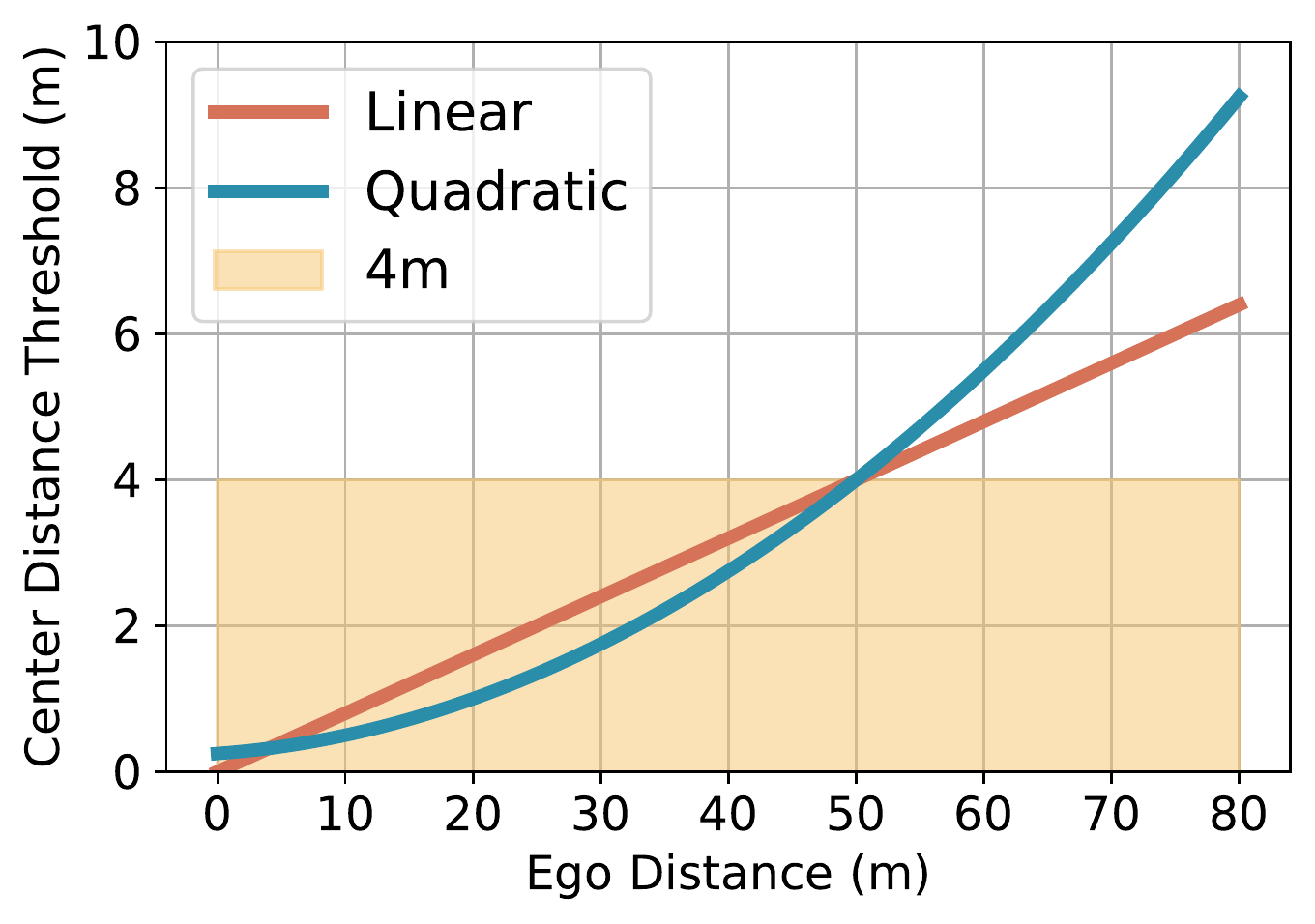}
\end{minipage} \hfill
\begin{minipage}[c]{0.25\textwidth}
\caption{\footnotesize
Standard metrics count positive detections using a fixed threshold (e.g., 4m). Our metrics are more reasonable that use adaptive thresholds that grow {\em linearly} or {\em quadratically} w.r.t depth. This imposes relaxed thresholds for far-field objects as humans cannot perceiving far localization too~\cite{welchman2004human,rushton2009observers} and strict thresholds for near-field objects.
Based on this figure, we set parameters in Eq.~\ref{eq:1}-\ref{eq:3}.
}
\label{fig:metric_thresh}
\end{minipage}
\vspace{-6mm}
\end{figure}

We design two metrics, {\tt linear} and {\tt quadratic}. 
The {\tt quadratic} distance-based threshold can be derived from standard error analysis of stereo triangulation. 
It is important to know that the distance-adaptive thresholds not only impose reasonable / relaxed thresholds for far-field objects, but also stricter ones in the near-field.

Mathematically, for {\tt linear}, we use 4m threshold at a distance of 50m and 0m threshold at 0m and derive the relation. We pick 4m for 50m based on the highest center-distance threshold provided by nuScenes on the range of 0-50m for cars, trucks and bus for standard evaluation.
\vspace{-2mm}
\begin{align}\small
\text{thresh}(d)  = d/12.5
\label{eq:1}
\end{align}
For {\tt quadratic} we use 4m threshold at 50m, 0.5m threshold at 10m and 1m at 20m. Again, we pick 0.5m and 1m at 10m and 20m distance respectively based on the lowest thresholds provided by nuScenes for standard evaluation.
\vspace{-2mm}
\begin{align}
\text{thresh}(d)  = 0.25 + 0.0125d + 0.00125(d^2)
\label{eq:2}
\end{align}

{\bf Elliptical metrics:} Till now, the metrics we have covered can be considered ``circular" in nature since they match predictions with the ground truth based on a circular matching criteria (IoU, center-distance, etc.). However, in a real world scenario, identifying the objects in the same lane can be considered more important, therefore, we also design an elliptical thresholding scheme that allows for larger longitudinal threshold (along the major axis $y$ of ego-vehicle) and smaller lateral threshold ($x$). The boundary of such an ellipse is given by
\begin{align}\small
\frac{312.5(x - x')^2 + 78.125(y - y')^2}{x'^2 + y'^2} = 1
\label{eq:3}
\end{align}
where (x, y) and (x$'$, y$'$)  denote distance of predicted and ground truth boxes from the ego vehicle in meters.

In principle, any 3D detector can be trained for detecting far-field 3D objects. But the core questions are how well they perform, what their limitations are, and how to improve their performance. The foremost goal of this work is to shed light on these questions by exploring various existing baselines for Far3Det under proposed evaluation settings.

\subsection{Multimodal Fusion for 3D Detection}

Multimodal detection is an active research field, where numerous methods are proposed in the literature. Existing multimodal detectors vary in terms of how to fuse multimodal information, e.g., on raw data, over features, or fusing single-modal detections. We extend the CLOCs baseline for 3D fusion. We also propose two late fusion algorithms namely distance fusion and AdaNMS fusion described in this section.

{\bf CLOCs3D.}
We extend CLOCs~\cite{pang2009clocs} to perform the fusion of 3D detections.
We modify the features to include the 3D IoU ($IoU_{\{i,j\}}^{3D}$), euclidean distance between candidates ($d_{ij}$), the distance from the ego vehicle ($d_j$) and prediction scores $s_{i}$ and $s_{j}$. Each element of the feature tensor can be calculated as $T_{i, j} =  \{IoU_{\{i,j\}}^{3D}, s_{i}^{3D}, s_{i}^{3D},d_{ij}, d_j\}$.

{\bf Distance-based Fusion}.
By analyzing the performance of single-modal and multimodal detectors on objects at different distance ranges, we find that the lidar-based detectors have a dominant performance on near-field objects. On the other hand, the image-based detectors perform better for detecting far-away objects with sparse lidar points.
Concretely, considering the operation distance range $d$, we have the final detections for class-$c$:
\begin{align}\small
{\cal D}_c = &{\cal D}_c^{(d<t_c)}  \  \cup \ {\cal D}_c^{(d>t_c)} \\
\text{where \ \ } &\left\{
        \begin{array}{ll}
             {\cal D}_c^{(d<t_c)}  \leftarrow \text{lidar-based detections}  \\
            {\cal D}_c^{(d>t_c)} \leftarrow \text{Fusion-based detections}
        \end{array}
        \right.
\end{align}

{\bf{Adaptive NMS (AdaNMS)}}. 
We notice that the far-field single-modal detections are noisy and often produce overlapping detections for the same ground-truth object.
To suppress such overlapping detections in the far field, we propose to use a smaller IoU threshold.
To this end, we introduce a distance adaptive IoU threshold for NMS, AdaNMS for short. To compute the adaptive threshold for an arbitrary distance, we qualitatively select two IoU thresholds that work sufficiently well on close range and far-field objects. For objects at given close-by distance \textit{$d_1$}, we pick an overlap threshold \textit{$c_1$} and pick threshold \textit{$c_2$} for far-field objects at distance \textit{$d_2$}. 
Our adaptive IoU threshold at an arbitrary distance \textit{d} is then given by:
\vspace{-2mm}
\begin{align}\small
\text{IoU}_{thresh} = (\textit{d} - \textit{$d_1$})\left(\frac{\textit{$c_2$} - \textit{$c_1$}}{\textit{$d_2$} - \textit{$d_1$}}\right) + c_1
\end{align}
 We pick distance ranges \textit{$d_1$} = 10m and \textit{$d_2$} = 70m and qualitatively select thresholds \textit{$c_1$} = 0.2 and \textit{$c_2$} = 0.05 respectively.

\section{Experiments}
\label{sec:exp}
We present the results for evaluating common detectors under both the standard and the proposed protocol. Next, we evaluate a suite of different 3D detectors, including both single- and multi-modal methods, under our finalized evaluation protocol. We evaluate the models with inclusion of unoccluded zero lidar point objects and we also report numbers on the elliptical thresholding metric.

{\setlength{\tabcolsep}{0.35em}
\begin{table}[t]
\footnotesize
\centering
\caption{\small 
Class-specific detection range used in the original nuScenes~\cite{Caesar_2020_CVPR} benchmark.}
\vspace{-3mm}
\begin{tabular}{c c c c c c c}
\toprule
\textbf{Class} & Car & Truck & Bus & Pedestrian &  Motorcycle & Traffic Cone \\ 
\cmidrule(r){1-1}  \cmidrule(r){2-2} \cmidrule(r){3-3}  \cmidrule(r){3-3}  \cmidrule(r){4-4}  \cmidrule(r){5-5} \cmidrule(r){6-6}  \cmidrule(r){7-7}
\textbf{Range (m)} & 50 & 50 & 50 & 40 & 40 & 30      \\ \hline
\end{tabular}
\label{tab:class-thresholds}
\vspace{-3mm}
\end{table}
}

\subsection{Setup}

We conduct our experiments on nuScenes~\cite{Caesar_2020_CVPR} and Far nuScenes (introduced in Sec~\ref{ssec:dataset_eval}). nuScenes is a multimodal 3D detection dataset containing synchronized captures of RGB, lidar, and RADAR sequences. The dataset contains 1000 scenes with around 6 hours of capture in total. The lidar sweeps are gathered at 20Hz and dense 3D bounding box annotations are provided at 2Hz. Such diverse well-organized data in the autonomous driving setting make it one of the most established benchmarks for 3D object detectors. Far nuScenes is a subset of nuScenes with high-quality annotations, especially for far-away objects. Since the literature of 3D detection mainly focuses on two modalities of lidar and RGB, we base our analysis on both these modalities.

\begin{table}[t]
\centering
\setlength{\tabcolsep}{0.15em}
\scriptsize
\caption{\small
Comparison of evaluation protocols (3D mAP) on nuScenes for  lidar-based~\cite{yin2020center} and image-based~\cite{wang2021fcos3d} detectors. We calculate the AP metric using the default thresholds of \{0.5m, 1m, 2m 4m\}, the proposed adaptive linear threshold and  quadratic threshold respectively. 
We find low numbers for image-based method in 50-80m range using default metric that are not consistent with the visualization shown in Fig.~\ref{fig:metric_visual} (see more visuals in the supplement). We posit this is due to a too strict distance tolerance (e.g. 0.5m) in far field.
}
\vspace{-3mm}
\begin{tabular}{c c ccc ccc} 
    \toprule
     & & \multicolumn{3}{c}{0-50m} & \multicolumn{3}{c}{50-80m} \\
    \cmidrule(r){3-5} \cmidrule(r){6-8}
    {Class} & {Method} & {Default} & {Linear}  & {Quadratic} & {Default} & {Linear}  & {Quadratic}  \\
    \midrule
    \multirow{2}{*}{Car} & lidar-based~\cite{yin2020center} & 86.0 & 94.9 & 94.1 & \textbf{10.4} & 22.0 & 22.3  \\ 
     & Image-based~\cite{wang2021fcos3d} & 44.3 & 69.4 & 46.5 & 4.6 & \textbf{38.0}  & \textbf{43.3}  \\ \hline

  \multirow{2}{*}{Truck} & lidar-based~\cite{yin2020center}  & 56.7 & 64.1 & 59.3 & \textbf{5.1} & 9.7 & 9.9   \\
   & Image-based~\cite{wang2021fcos3d} & 22.5 & 31.7 & 17.9 & 3.2 & \textbf{13.6} &  \textbf{17.4}  \\  \hline

   \multirow{2}{*}{Pedestrian} & lidar-based~\cite{yin2020center} & 88.1 & 88.5 & 87.7 & \textbf{13.4} & \textbf{14.0} & \textbf{14.0}  \\ 
   & Image-based~\cite{wang2021fcos3d} & 32.7 & 38.0 & 21.5 & 2.6 & 9.8 & 12.0 \\ 
    \bottomrule
\end{tabular}
\vspace{-4mm}
\label{tab:adaptive_threshold_all_nuScenes}
\end{table}

Note that the original nuScenes benchmark defines (short) per-class ranges for evaluating objects of different classes (shown in Table~\ref{tab:class-thresholds}), presumably because different classes have different prior object shape/size such as pedestrians are small therefore lidar might not return points on them at distance.
Following the discussion in Sec~\ref{sec:intro}, we argue that such maximum distances are insufficient for the application of medium- or high-speed driving. Therefore, we increase the detection range to 80 meters for all classes for both evaluation and training. We train the models on this updated setup and unless otherwise stated, we use models trained with this setup for evaluation purposes.

We adopt two popular 3D detectors, CenterPoint~\cite{yin2020center} and FCOS3D~\cite{wang2021fcos3d}, as representative lidar- and image-based detectors respectively.  FCOS3D achieves the state-of-the-art image-based 3D detection performance on multiple leaderboards. Therefore, we use FCOS3D in our work to study Far3Det. A lidar-based 3D detector takes as input an aggregation of lidar sweeps and outputs detections (cuboids coordinates and class labels). This type of detector greatly outperforms image-based detectors in various 3D detection benchmarks~\cite{Geiger2012CVPR,Caesar_2020_CVPR}. We show it is not true in terms of Far3Det. Among numerous lidar-based 3D detectors, we choose CenterPoint~\cite{yin2020center} because it achieves the state-of-the-art 3D detection performance on various benchmarks~\cite{sun2020scalability,Caesar_2020_CVPR}. We later introduce additional baselines for a more comprehensive evaluation.

Unless otherwise stated, we adopt the popular codebase MMDetection3D toobox~\cite{mmdet3d2020} for baseline methods. We adopt default hyperparameters except for (a) the learning rate, since we are training with 4 GPUs, half of the standard setup and a smaller learning rate should be used under this settings \cite{goyal2017accurate}, (b) the minimum number of lidar points in a box to 1 (from the default value of 5) to allow sparse detections, and (c) the point cloud range [-80, -80, -5, 80, 80, 3] to include the distant objects detection. Specifically, notable hyperparameters for CenterPoint are AdamW optimizer with cyclic learning rate scheduler, voxel size of [0.075, 0.075, 0.2]; for FCOS3D are SGD optimizer with 0.9 momentum, 12 epochs and image resolution 1600x900. We use circle-NMS, double flip to achieve higher accuracy for CenterPoint. We follow the standard procedure to train the FCOS3D model with the initial depth weight set to 0.2 and then fine-tune the model with depth weight 1.0. Since MVP~\cite{yin2021multimodal} trains the CenterPoint on densified point clouds using virtual points, we use the same setting to train it.

\begin{table}[t]
\centering
\setlength{\tabcolsep}{0.15em}
\small
\caption{\small
Far nuScenes version of Table~\ref{tab:adaptive_threshold_all_nuScenes}.
We observe similar trend as in Table~\ref{tab:adaptive_threshold_all_nuScenes} but higher numbers, which we believe 
realistically reflect the 3D detection performance in the far-field.  
}
\scriptsize
\vspace{-3mm}
\begin{tabular}{c c ccc ccc} 
    \toprule
     & & \multicolumn{3}{c}{0-50m} & \multicolumn{3}{c}{50-80m} \\
    \cmidrule(r){3-5} \cmidrule(r){6-8}
    {Class} & {Method} & {Default} & {Linear}  & {Quadratic} & {Default} & {Linear}  & {Quadratic}  \\
    \midrule
    \multirow{2}{*}{Car} & lidar-based~\cite{yin2020center} & 91.2 & 94.9 & 94.1 & 19.2 & 28.5 & 29.1  \\ 
     & Image-based~\cite{wang2021fcos3d} & 57.3 & 72.1 & 49.5 & 11.9 & 45.9 & 51.7  \\ \hline
  \multirow{2}{*}{Truck} & lidar-based~\cite{yin2020center}  & 62.8 & 66.3 & 59.2 & 6.5 & 12.9 & 13.4   \\ 
   & Image-based~\cite{wang2021fcos3d} & 28.6 & 28.2 & 14.9 & 3.2 & 13.4 & 18.0   \\  \hline

   \multirow{2}{*}{Pedestrian} & lidar-based~\cite{yin2020center} & 93.0 & 92.9 & 91.9 & 16.9 & 17.4 & 17.5  \\ 
   & Image-based~\cite{wang2021fcos3d} & 43.1 & 42.7 & 25.6 & 4.4 & 16.6 & 20.5 \\ 
    \bottomrule
\end{tabular}
\label{tab:adaptive_threshold_far_nuScenes}
\vspace{-4mm}
\end{table}

\subsection{Results for Proposed Evaluation Protocol}

{\bf Metrics for Far3Det Evaluation.}
We evaluate the car, truck, and pedestrian mAP of lidar- (CenterPoint~\cite{yin2020center}) and image-based (FCOS3D~\cite{wang2021fcos3d}) models on 0-50m and 50-80m distance range. We use CP as an abbreviation for CenterPoint. Table \ref{tab:adaptive_threshold_all_nuScenes} shows the mAP values for the 0-50m and 50-80m range on nuScenes validation set (other classes in appendix). We observe that the Far3Det mAP is lower for image-based method (column d) compared to lidar-based method when we use the default nuScenes thresholding scheme, however when we use our proposed linear and quadratic thresholding schemes (d \& e), we observe that image-based method outperforms lidar-based method for cars and trucks. In the next section, we perform similar analysis on Far nuScenes.

{\bf Evaluation on Far nuScenes.} 
Based on our observation of missing annotations in nuScenes (Fig. \ref{fig:annotation_stats}), we again perform same evaluation as Table \ref{tab:adaptive_threshold_all_nuScenes} on the Far nuScenes to get more reliable numbers. Table \ref{tab:adaptive_threshold_far_nuScenes} shows that the corresponding 3D mAP values are higher on Far nuScenes compared to the nuScenes for 50-80m range. On visual examination of random samples, we observed that the models were able to generate the predictions for far field but since the ground truth was missing for them in nuScenes, we get lower mAP values compared to Far nuScenes. We also observed that mAP of lidar-based method does not increase in same proportion as that of image-based method as the distance threshold increases. 
This can be attributed to the fact that the lidar-based methods predict accurate boxes if they get enough lidar points but since the lidar suffer from sparsity problem at distance, the number of detections are quite small. Image-based methods suffer relatively less from this problem, hence, their accuracy drop is comparatively less than its lidar counterpart.

We choose the parameters of our {\tt linear} and {\tt quadratic} adaptation using predefined thresholds at certain distances. We believe that this metric can be further improved by using certain heuristics based on the camera intrinsics.
Unless otherwise stated, we use Far nuScenes and {\tt linear} as our default evaluation protocol.

{\setlength{\tabcolsep}{0.08em}
\begin{table}[t]
\centering
\scriptsize
\caption{\small
Quantitative evaluation (3D mAP) on Far nuScenes under our proposed metrics based on linearly-adaptive distance thresholds.
First, we notice that all lidar-based detectors perform well for the near field but suffer greatly in the far-field. 
The VoxelNet-backbone CP (CenterPoint) significantly outperforms other detectors. The image-based detector FOCS3D significantly outperforms CP for far-field. All fusion methods are able to take the ``best of both worlds'', resulting in a significant gain for far-field (50-80m) accuracy. While being much simpler, our proposed methods NMS and AdaNMS fusion significantly improve upon more complicated baselines for all classes except Pedestrian, where a lower overlap threshold for NMS on far-field hurts the recall for cluttered scenes. *CP-VoxelNet is the same as CP appearing elsewhere in this paper. **CLOCs3D is an extension of CLOCs~\cite{pang2009clocs}. AdaNMS has two versions, one trained with MVP.}
\vspace{-3mm}
\begin{tabular}{c  cc cc cc cc} 
    \toprule
     & \multicolumn{2}{c}{Modality} & \multicolumn{2}{c}{Car} & \multicolumn{2}{c}{Truck} & \multicolumn{2}{c}{Pedestrian} \ \\
    \cmidrule(r){2-3} \cmidrule(r){4-5} \cmidrule(r){6-7} \cmidrule(r){8-9}
    {Model} & {lidar} & {Cam}  & {0-50m} & {50-80m} & {0-50m}  & {50-80m} & {0-50m} & {50-80m} \\
    \midrule
      CP-VoxelNet$^{*}$~\cite{yin2020center} & \cmark &  & 94.9 & 28.5 & 66.3 & 12.9 & 92.9 & 17.4 \\
CP-PointPillars~\cite{yin2020center}  & \cmark &  & 92.7 & 14.4 & 56.3 & 3.1 & 85.2 & 6.7 \\ 
PointPillars-FPN~\cite{radosavovic2020designing} & \cmark & & 89.7 & 7.8 & 48.0 & 0.7 & 83.3 & 1.9 \\ 
PointPillars-SECFPN~\cite{radosavovic2020designing} & \cmark & & 88.2 & 5.9 & 54.8 & 1.4 & 83.4 & 1.2 \\ 
SSN-SECFPN~\cite{zhu2020ssn} & \cmark &  & 89.8 & 8.3 & 52.7 & 1.2 & 75.7 & 1.1 \\ 
\Xhline{2\arrayrulewidth}
FCOS3D~\cite{wang2021fcos3d} &  & \cmark & 72.1 & 45.9 & 28.2 & 13.4 & 42.7 & 16.6 \\
\Xhline{2\arrayrulewidth}
Bayesian Fusion~\cite{chen2021multimodal} & \cmark & \cmark & 94.3 & 54.8 & 62.3 & 21.9 & 93.1 & 24.0 \\ 
CLOCs3D$^{**}$ & \cmark & \cmark &  94.3  & 54.8 & 62.3 & 21.9 & 93.1 & 24.0 \\ 
NMS Fusion & \cmark & \cmark & 94.9 & 55.1 & 66.3 & 24.2 & 92.9 & 25.6 \\ 

AdaNMS (CP, FCOS3D) & \cmark & \cmark & 94.9 & 57.2 & 66.3 & 25.2 & 92.9 & 21.3 \\ 
MVP~\cite{yin2021multimodal} & \cmark & \cmark & \textbf{95.9} & 70.6 & \textbf{70.0} & 46.7 & \textbf{96.1} & \textbf{58.9} \\ 
AdaNMS (MVP, FCOS3D) & \cmark & \cmark & \textbf{95.9} & \textbf{72.8} & \textbf{70.0} & \textbf{49.3} & \textbf{96.1} & 57.6 \\ 
    \bottomrule
\end{tabular}
\label{tab:verified_nuscenes}
\vspace{-2mm}
\end{table}
}

\begin{table}[t]
\centering
\setlength{\tabcolsep}{0.08em}
\scriptsize
\caption{\small 
AP computed with quadratically-growing threshold (as shown in Table~\ref{tab:verified_nuscenes}).
Overall, results and trends are qualitatively similar to those for a linearly-growing threshold.
}
\vspace{-3mm}
\begin{tabular}{c  cc cc cc cc} 
    \toprule
     & \multicolumn{2}{c}{Modality} & \multicolumn{2}{c}{Car} & \multicolumn{2}{c}{Truck} & \multicolumn{2}{c}{Pedestrian} \ \\
    \cmidrule(r){2-3} \cmidrule(r){4-5} \cmidrule(r){6-7} \cmidrule(r){8-9}
    {Model} & {lidar} & {Cam}  & {0-50m} & {50-80m} & {0-50m}  & {50-80m} & {0-50m} & {50-80m} \\
    \midrule
  CP-VoxelNet$^{*}$~\cite{yin2020center} & \cmark &  & \textbf{94.1} & 29.1 & \textbf{59.2} & 13.4 & 91.9 & 17.5 \\ 
CP-PointPillars~\cite{yin2020center}  & \cmark &  & 91.6 & 14.8 & 53.1 & 3.2 & 83.7 & 6.7 \\ 
PointPillars-FPN~\cite{radosavovic2020designing} & \cmark & & 88.3 & 8.1 & 41.1 & 0.7 & 82.6 & 2.0 \\ 
PointPillars-SECFPN~\cite{radosavovic2020designing} & \cmark & & 86.5 & 6.3 & 51.2 & 1.4 & 77.9 & 1.2\\ 
SSN-SECFPN~\cite{zhu2020ssn} & \cmark &  & 88.3 & 8.6 & 46.9 & 1.2 & 74.7 & 1.1\\ 
\Xhline{2\arrayrulewidth}
FCOS3D~\cite{wang2021fcos3d} &  & \cmark & 49.5 & 51.7 & 14.9 & 18.1 & 25.7 & 20.5 \\ 
\Xhline{2\arrayrulewidth}
Bayesian Fusion~\cite{chen2021multimodal} & \cmark & \cmark & 93.5 & 57.8 & 56.4 & 24.5 & 92.2 & 25.4 \\ 
CLOCs3D$^{**}$ & \cmark & \cmark &  93.5  & 57.8 & 56.4 & 24.5 & 92.2 & 25.4 \\ 
NMS Fusion (Ours) & \cmark & \cmark & 94.1 & 58.1 & 59.2 & 27.4 & 91.9 & 27.1 \\ 
AdaNMS Fusion (Ours) & \cmark & \cmark & 94.1 & 60.5 & 59.2 & 28.1 &  91.9 & 22.6 \\
MVP~\cite{yin2021multimodal} & \cmark & \cmark & \textbf{95.17} & 72.10 & \textbf{94.44} & 48.25 & \textbf{94.44} & \textbf{59.55} \\ 
AdaNMS (MVP, FCOS3D) & \cmark & \cmark & \textbf{95.17} & \textbf{72.89} & \textbf{94.44} & \textbf{49.31} & \textbf{94.44} & 57.57 \\ 
\bottomrule
\end{tabular}
\vspace{-4mm}
\label{tab:verified_nuScenes_q}
\end{table}

\subsection{Results}

{\bf Single Modality Baselines.}
Since lidar-based methods significantly outperform image-based methods on the leaderboards, we train five lidar-based models to demonstrate that none of then is better than the image-based method for detecting distant objects. We use the current state-of-the-art CenterPoint (both Pointpillar and Voxelnet based) \cite{yin2020center}, Pointpillar FPN, Pointpillar-based RegNetX \cite{radosavovic2020designing}, and Shape Signature Networks (SSN)~\cite{zhu2020ssn}. For the image-based method, we use the FCOS3D~\cite{wang2021fcos3d}.

Table \ref{tab:verified_nuscenes} summarizes the 3D mAP values for car, truck and pedestrian. We train our model on all the classes present in nuScenes dataset but we show only these three classes (see results on other classes in appendix). We observe that for 0-50m all the lidar-based methods significantly outperform the image-based method (FCOS3D). For the far-field (50-80m), all lidar-based methods experience a sharp drop in accuracy for all the classes. We observe that the image-based method outperforms all lidar-based methods in this range for the car and truck class and has comparable performance for the pedestrian class.

{\bf Multimodal Fusion Baselines.}
We fuse CenterPoint~\cite{yin2020center} VoxelNet) with FCOS3D~\cite{wang2021fcos3d} using various methods- NMS Fusion, AdaNMS Fusion, and Learning-based Fusion CLOCs3D. Since, the MVP~\cite{yin2021multimodal} can also be viewed as a lidar-centric fusion (it trains CenterPoint with virtual points densified using 2D image segmentation), we also fuse it with FCOS3D model. For all the fusion-based methods, we use the unprocessed detections from both the single modality detectors to fuse them together.

{
\setlength{\tabcolsep}{1.25em} 
\begin{table}[t]
\centering
\scriptsize
\caption{\small
We include the unoccluded objects with zero lidar points using strategy described in Section \ref{subsec: zero_points} and calculate the mAP for 50-80m distance range. Clearly, our AdaNMS (MVP+FCOS3D) outperforms others on all categories except the Pedestrian category. We see a slight decline in performance on this class as a distance adaptive IOU hurts the recall in a cluttered scene.
}
\vspace{-3mm}
\begin{tabular}{c ccc} 
    \toprule
     & \multicolumn{3}{c}{50-80m} \\
    \cmidrule(r){2-4} 
    {Method} & {Car  } & {Truck  }  & {Pedestrian  }\\
    \midrule
        CP~\cite{yin2020center} & 20.5 & 8.6 & 11.2   \\ 
        FCOS3D~\cite{wang2021fcos3d} & 35.2 & 9.7 & 11.7 \\ 
        AdaNMS (CP, FCOS3D) & 45.2 & 18.5 & 14.4 \\ 
        MVP~\cite{yin2021multimodal} & 55.2 & 35.5 & \textbf{44.2} \\ 
        AdaNMS (MVP, FCOS3D) & \textbf{60.3} & \textbf{36.0} & 42.6 \\  
    
    \bottomrule
\end{tabular}
\label{tab:occluded_objects}
\vspace{-5mm}
\end{table}
}

{\bf Multimodal Fusion Results.} 
Table \ref{tab:verified_nuscenes} provides the detailed comparison of various fusion baselines.. We observe that all the fusion methods outperform the single modality methods on far-field detection.Note that MVP has superior performance compared to all other methods and our AdaNMS (MVP, FCOS3D) fusion works the best for detection of distant objects on Far nuScenes. Thus we can conclude that our late-fusion strategy is able to combine the predictions of both models and generate more accurate predictions for far-field objects. Specifically, we observe that the mAP of car increases by 11.3, truck by 10.8 over FCOS3D for AdNMS fusion.

\subsection{Detection of Zero-lidar Point Objects}
\label{subsec: zero_points}

As explained in Sec.~\ref{subsec: zero_points_long}, we include the zero lidar point annotations in the evaluation to compare the methods. 
Table~\ref{tab:occluded_objects} summarizes the results for various methods on this dataset. Our AdaNMS achieves SOTA for most classes (Car, Truck, others in suppl) on this dataset. 
Also note that the mAP of all methods drops compared to corresponding values in Table~\ref{tab:verified_nuscenes}. This is expected since detecting these zero-lidar objects is hard even for image-based methods due to low visibility. In an attempt to align training and validation dataset, we also used similar approach to include the zero-lidar annotations in the training set. However, on re-training using this dataset, we didn't observe much gains. We attribute this to that the number of annotations added by this strategy are $\leq 5\%$ of the total dataset.

\subsection{Analysis using Adaptive Elliptical Metric}
As discussed in Sec~\ref{sec:metric}, we also evaluate the results using adaptive distance-based elliptical threshold boundary (Eq.~\ref{eq:3}) to penalize cross-lanes error more than same lane errors. Table~\ref{tab:elliptical metric} shows the 3D mAP values for car class using the circular and elliptical thresholding schemes. We observe that image-based model(FCOS3D) has higher drop in mAP compared to lidar-based model(CenterPoint). We believe that this is due to noisy image-based detections compared to lidar-based detections in the lateral direction as well (not only in longitudinal). As a result, decreasing the threshold in lateral direction to compensate for increase in the longitudinal direction impacts the accuracy of image-based methods. We believe that this can be explained by that fact that there is a higher exclusion of close-by region from elliptical boundary as compared to the circular boundary.

{\setlength{\tabcolsep}{0.56em}
\begin{table}[t]
\centering
\scriptsize
\caption{3D mAP values for car class using elliptical thresholding scheme proposed in Sec.~\ref{sec:metric}. We notice that the Elliptical metric and Linear metric produce same rankings of methods. The performance drop in image-based (FCOS3D) is much higher than that of lidar-based (CenterPoint) method when we calculate error based on the elliptical scheme. This can be due to exclusion of close-by region in elliptical thresholding scheme compared to the circular thresholding scheme.}
\vspace{-3mm}
\begin{tabular}{c  cc cc} 
    \toprule
     & \multicolumn{2}{c}{0-50m} & \multicolumn{2}{c}{50-80m} \\
    \cmidrule(r){2-3} \cmidrule(r){4-5}
    {Method} & {Circular  } & {Elliptical  }  & {Circular  } & {Elliptical  }\\
    \midrule
    CP~\cite{yin2020center} & 94.9 & 94.6 & 28.5 & 28.5   \\ 
FCOS3D~\cite{wang2021fcos3d} & 72.2 & 62.4 & 45.9 & 38.8 \\
AdaNMS (CP, FCOS3D) & 94.9 & 94.6 & 57.2 & 54.2  \\
MVP~\cite{yin2021multimodal}  & 95.9 & 95.4 & 70.6 & 70.6 \\ 
AdaNMS (MVP, FCOS3D) & 95.9 & 95.5 & 72.8 & 71.0  \\
    \bottomrule
\end{tabular}
\vspace{-5mm}
\label{tab:elliptical metric}
\end{table}
}

\section{Conclusion}
\label{conclusion}

We highlight the problem of far-field 3D object detection (Far3Det) which is currently underexplored in the contemporary AV benchmarks. We provide a manually cleaned Far nuScenes dataset for evaluating the Far3Det models. We propose various adaptive distance-based thresholding schemes for calculating 3D mAP as the evaluation metric. 
We show that using RGB boosts Far3Det. We introduce simple yet effective fusion methods based on NMS, outperforming state-of-the-art lidar-based detectors for Far3Det.

{\small \noindent{\bf Acknowledgements}. This work was supported by the CMU Argo AI Center for Autonomous Vehicle Research.}



{\small
\bibliographystyle{splncs04}
\bibliography{arxiv}

\begin{thebibliography}{10}
\providecommand{\url}[1]{\texttt{#1}}
\providecommand{\urlprefix}{URL }
\providecommand{\doi}[1]{https://doi.org/#1}

\bibitem{nuscenes:forum}
nuScenes authors: 0 lidar+radar points.
  \url{https://forum.nuscenes.org/t/0-lidar-radar-points/650/4} (2021)

\bibitem{Caesar_2020_CVPR}
Caesar, H., Bankiti, V., Lang, A.H., Vora, S., Liong, V.E., Xu, Q., Krishnan,
  A., Pan, Y., Baldan, G., Beijbom, O.: nuscenes: A multimodal dataset for
  autonomous driving. In: Proceedings of the IEEE/CVF Conference on Computer
  Vision and Pattern Recognition (CVPR) (June 2020)

\bibitem{chen2017multiview}
Chen, X., Ma, H., Wan, J., Li, B., Xia, T.: Multi-view 3d object detection
  network for autonomous driving. In: Proceedings of the IEEE conference on
  Computer Vision and Pattern Recognition. pp. 1907--1915 (2017)

\bibitem{chen2021multimodal}
Chen, Y.T., Shi, J., Mertz, C., Ramanan, D., Kong, S.: Multimodal object
  detection via probabilistic ensembling. In: European Conference on Computer
  Vision (ECCV) (2022)

\bibitem{mmdet3d2020}
Contributors, M.: {MMDetection3D: OpenMMLab} next-generation platform for
  general {3D} object detection.
  \url{https://github.com/open-mmlab/mmdetection3d} (2020)

\bibitem{doermann2000tools}
Doermann, D., Mihalcik, D.: Tools and techniques for video performance
  evaluation. In: ICPR (2000)

\bibitem{Geiger2012CVPR}
Geiger, A., Lenz, P., Urtasun, R.: Are we ready for autonomous driving? the
  kitti vision benchmark suite. In: Conference on Computer Vision and Pattern
  Recognition (CVPR) (2012)

\bibitem{goyal2017accurate}
Goyal, P., Doll{\'a}r, P., Girshick, R., Noordhuis, P., Wesolowski, L., Kyrola,
  A., Tulloch, A., Jia, Y., He, K.: Accurate, large minibatch sgd: Training
  imagenet in 1 hour. arXiv preprint arXiv:1706.02677  (2017)

\bibitem{greibe2007braking}
Greibe, P.: Braking distance, friction and behaviour. Trafitec, Scion-DTU
  (2007)

\bibitem{henrion2013qualitative}
Henrion, M., Druzdzel, M.J.: Qualitative propagation and scenario-based
  explanation of probabilistic reasoning (2013)

\bibitem{hu2019wysiwyg}
Hu, P., Ziglar, J., Held, D., Ramanan, D.: What you see is what you get:
  Exploiting visibility for 3d object detection. In: Proceedings of the
  IEEE/CVF Conference on Computer Vision and Pattern Recognition (2020)

\bibitem{hung2022let}
Hung, W.C., Kretzschmar, H., Casser, V., Hwang, J.J., Anguelov, D.: Let-3d-ap:
  Longitudinal error tolerant 3d average precision for camera-only 3d
  detection. arXiv preprint arXiv:2206.07705  (2022)

\bibitem{joshua1992mitigation}
Joshua, S.C., Saka, A.A.: Mitigation of sight-distance problem for unprotected
  left-turning traffic at intersections. No.~1356 (1992)

\bibitem{king2017top}
King, L.: Top 15 causes of car accidents and how you can prevent them (2017)

\bibitem{ku2018joint}
Ku, J., Mozifian, M., Lee, J., Harakeh, A., Waslander, S.: Joint 3d proposal
  generation and object detection from view aggregation. IROS  (2018)

\bibitem{lang2019pointpillars}
Lang, A.H., Vora, S., Caesar, H., Zhou, L., Yang, J., Beijbom, O.:
  Pointpillars: Fast encoders for object detection from point clouds. In:
  {IEEE} Conference on Computer Vision and Pattern Recognition, {CVPR}. pp.
  12697--12705. Computer Vision Foundation / {IEEE} (2019)

\bibitem{liang2018deep}
Liang, M., Yang, B., Wang, S., Urtasun, R.: Deep continuous fusion for
  multi-sensor 3d object detection. In: Proceedings of the European conference
  on computer vision (ECCV). pp. 641--656 (2018)

\bibitem{meyer2019sensor}
Meyer, G.P., Charland, J., Hegde, D., Laddha, A., Vallespi-Gonzalez, C.: Sensor
  fusion for joint 3d object detection and semantic segmentation. In:
  Proceedings of the IEEE/CVF Conference on Computer Vision and Pattern
  Recognition Workshops. pp.~0--0 (2019)

\bibitem{nowak2010reliable}
Nowak, S., R{\"u}ger, S.: How reliable are annotations via crowdsourcing: a
  study about inter-annotator agreement for multi-label image annotation. In:
  Proceedings of the international conference on Multimedia information
  retrieval. pp. 557--566 (2010)

\bibitem{pang2009clocs}
Pang, S., Morris, D., Radha, H.: Clocs: Camera-lidar object candidates fusion
  for 3d object detection  (2020)

\bibitem{peri2022towards}
Peri, N., Dave, A., Ramanan, D., Kong, S.: Towards long-tailed 3d detection.
  In: Conference on Robot Learning (CoRL) (2022)

\bibitem{qi2018frustum}
Qi, C.R., Liu, W., Wu, C., Su, H., Guibas, L.J.: Frustum pointnets for 3d
  object detection from rgb-d data. In: Proceedings of the IEEE conference on
  computer vision and pattern recognition. pp. 918--927 (2018)

\bibitem{radosavovic2020designing}
Radosavovic, I., Kosaraju, R.P., Girshick, R., He, K., Dollár, P.: Designing
  network design spaces. In: CVPR (2020)

\bibitem{rushton2009observers}
Rushton, S.K., Duke, P.A.: Observers cannot accurately estimate the speed of an
  approaching object in flight. Vision research  \textbf{49}(15),  1919--1928
  (2009)

\bibitem{sindagi2019mvxnet}
Sindagi, V.A., Zhou, Y., Tuzel, O.: Mvx-net: Multimodal voxelnet for 3d object
  detection. In: 2019 International Conference on Robotics and Automation
  (ICRA). pp. 7276--7282. IEEE (2019)

\bibitem{sun2020scalability}
Sun, P., Kretzschmar, H., Dotiwalla, X., Chouard, A., Patnaik, V., Tsui, P.,
  Guo, J., Zhou, Y., Chai, Y., Caine, B., et~al.: Scalability in perception for
  autonomous driving: Waymo open dataset. In: Proceedings of the IEEE/CVF
  Conference on Computer Vision and Pattern Recognition. pp. 2446--2454 (2020)

\bibitem{vanderbilt2009traffic}
Vanderbilt, T.: Traffic: Why we drive the way we do (and what it says about
  us). Vintage (2009)

\bibitem{vondrick2013efficiently}
Vondrick, C., Patterson, D., Ramanan, D.: Efficiently scaling up crowdsourced
  video annotation. International journal of computer vision  \textbf{101}(1),
  184--204 (2013)

\bibitem{vora2020pointpainting}
Vora, S., Lang, A.H., Helou, B., Beijbom, O.: Pointpainting: Sequential fusion
  for 3d object detection. In: Proceedings of the IEEE/CVF conference on
  computer vision and pattern recognition. pp. 4604--4612 (2020)

\bibitem{wang2019exploit}
Wang, G., Wang, Y., Zhang, H., Gu, R., Hwang, J.N.: Exploit the connectivity:
  Multi-object tracking with trackletnet. In: Proceedings of the 27th ACM
  International Conference on Multimedia. pp. 482--490 (2019)

\bibitem{wang2021fcos3d}
Wang, T., Zhu, X., Pang, J., Lin, D.: Fcos3d: Fully convolutional one-stage
  monocular 3d object detection. In: Proceedings of the IEEE/CVF International
  Conference on Computer Vision. pp. 913--922 (2021)

\bibitem{wang2019pseudo}
Wang, Y., Chao, W.L., Garg, D., Hariharan, B., Campbell, M., Weinberger, K.Q.:
  Pseudo-lidar from visual depth estimation: Bridging the gap in 3d object
  detection for autonomous driving. In: Proceedings of the IEEE/CVF Conference
  on Computer Vision and Pattern Recognition. pp. 8445--8453 (2019)

\bibitem{8968513}
Wang, Z., Jia, K.: Frustum convnet: Sliding frustums to aggregate local
  point-wise features for amodal 3d object detection. In: 2019 IEEE/RSJ
  International Conference on Intelligent Robots and Systems (IROS). pp.
  1742--1749 (2019). \doi{10.1109/IROS40897.2019.8968513}

\bibitem{welchman2004human}
Welchman, A.E., Tuck, V.L., Harris, J.M.: Human observers are biased in judging
  the angular approach of a projectile. Vision research  \textbf{44}(17),
  2027--2042 (2004)

\bibitem{wilson2021argoverse}
Wilson, B., Qi, W., Agarwal, T., Lambert, J., Singh, J., Khandelwal, S., Pan,
  B., Kumar, R., Hartnett, A., Pontes, J.K., Ramanan, D., Carr, P., Hays, J.:
  Argoverse 2: Next generation datasets for self-driving perception and
  forecasting. In: Thirty-fifth Conference on Neural Information Processing
  Systems Datasets and Benchmarks Track (Round 2) (2021),
  \url{https://openreview.net/forum?id=vKQGe36av4k}

\bibitem{xu2018pointfusion}
Xu, D., Anguelov, D., Jain, A.: Pointfusion: Deep sensor fusion for 3d bounding
  box estimation. In: Proceedings of the IEEE conference on computer vision and
  pattern recognition. pp. 244--253 (2018)

\bibitem{xu2021fusionpainting}
Xu, S., Zhou, D., Fang, J., Yin, J., Bin, Z., Zhang, L.: Fusionpainting:
  Multimodal fusion with adaptive attention for 3d object detection. In: 2021
  IEEE International Intelligent Transportation Systems Conference (ITSC). pp.
  3047--3054. IEEE (2021)

\bibitem{yan2018second}
Yan, Y., Mao, Y., Li, B.: Second: Sparsely embedded convolutional detection.
  Sensors  \textbf{18}(10), ~3337 (2018)

\bibitem{yin2020center}
Yin, T., Zhou, X., Kr{\"a}henb{\"u}hl, P.: Center-based 3d object detection and
  tracking. In: CVPR (2021)

\bibitem{yin2021multimodal}
Yin, T., Zhou, X., Kr{\"a}henb{\"u}hl, P.: Multimodal virtual point 3d
  detection. NeurIPS  (2021)

\bibitem{you2020pseudolidar}
You, Y., Wang, Y., Chao, W.L., Garg, D., Pleiss, G., Hariharan, B., Campbell,
  M., Weinberger, K.Q.: Pseudo-lidar++: Accurate depth for 3d object detection
  in autonomous driving. arXiv arXiv:1906.06310  (2020)

\bibitem{zhu2019class}
Zhu, B., Jiang, Z., Zhou, X., Li, Z., Yu, G.: Class-balanced grouping and
  sampling for point cloud 3d object detection. arXiv preprint arXiv:1908.09492
   (2019)

\bibitem{zhu2020ssn}
Zhu, X., Ma, Y., Wang, T., Xu, Y., Shi, J., Lin, D.: Ssn: Shape signature
  networks for multi-class object detection from point clouds. In: ECCV (2020)

\end{thebibliography}
}

\newpage

\section*{}
\begin{center}
{\bf \large Appendix}
\end{center}

{\em 

In this document, we first include detailed studies w.r.t different evaluation metrics to supplement Table 5 of the main paper.
Then, we include open-source code for far-field 3D detection (Far3Det) benchmarking, including Far nuScenes frame ids, metrics, and different fusion baselines (NMS, AdaNMS, and CLOCs3D).
Lastly, we attach a video demo to demonstrate the results of our lidar-based detector (CenterPoint~\cite{yin2020center}), image-based detector (FCOS3D~\cite{wang2021fcos3d}), and our NMS-based late-fused detector.
}

\section{Far Field Annotations}
We analyze the total number of far-field annotations per class as shown in Table~\ref{tab:anno_count}.

{
\setlength{\tabcolsep}{0.5em} 
\begin{table*}[t]
\centering
\small
\caption{\small
Class-specific far-field annotations (50-80m) in the nuScenes~\cite{Caesar_2020_CVPR} dataset. Here ``Construct" and ``Cone" are short for ``Construction Vehicle" and ``Traffic Cone" respectively. We perform this analysis for 
nuScenes Training set, nuScenes Validation set, and Far nuScenes. In general, there are fewer annotations for small objects (``Motorcycle'', ``Bicycle'', ``Traffic Cone'') than big ones (``Car'', ``Truck'', ``Bus'').
}
\vspace{-2mm}
\begin{tabular}{c cccccccccc} 
    \toprule
     & \multicolumn{10}{c}{50-80m} \\
    \cmidrule(r){2-11} 
    {Dataset} & {Car} & {Truck} & {Bus} & {Trailer} & {Construct}  & {Pedestrian} & {Motorcycle} & {Bicycle}  & {Cone} & {Barrier}\\
    \midrule
\textbf{Training} & 49758 & 15443 & 3751 & 5072 & 2345 &  16106 & 508 & 334 & 1424 & 5382  \\ \hline
\textbf{Validation} & 10435 & 3541 & 673 & 1084 & 740 & 3339 & 123 & 85 & 169 & 1469 \\ \hline
\textbf{Far nuScenes (Validation)} & 5043 & 1791 & 184 & 743 & 401 & 1427 & 9 & 35 & 81 & 765 \\ \hline
\end{tabular}
\label{tab:anno_count}
\end{table*}
}

\section{Evaluation Metric}

We evaluate all the baseline models and our proposed NMS and AdaNMS fusion methods for various thresholding schemes. For nuScenes default metric ( average of AP at {0.5, 1, 2 and 4} meters), we can observe that the lidar-based method (CenterPoint-VoxelNet) has higher AP compared to the image-based method (FCOS3D), (16.5 vs 11.5) for distant objects (50-80m) in contrast to what we observed at our proposed linear adaptive threshold-based metric. This occurs due the higher noise in the models prediction as the distance from the ego-vehicle increases.  Table \ref{tab:verified_nuScenes_4m} shows the performance of the various models at the evaluated on the 4m threshold. We can observe that this thresholding scheme follows the similar trend as our proposed linear and qudratic adaptive thresholding scheme.

{\setlength{\tabcolsep}{1.2em}
\begin{table*}[t]
\centering
\footnotesize
\caption{AP computed with 4m threshold. All numbers improve compared to the default nuScenes metric that averages AP over thresholds {\em up} to 4m.
Here, image-based far-field 3D detections are {\em far} more accurate than lidar detections (37 vs 28.1 for distant Cars from 50-80m). Fusion once again further improves results. Similar trends hold for Truck and Pedestrian. Overall, results and trends are qualitatively similar to those for a linearly-growing threshold (the default used in our paper).}
\vspace{-2mm}
\begin{tabular}{c  cc cc cc cc} 
    \toprule
     & \multicolumn{2}{c}{Modality} & \multicolumn{2}{c}{Car} & \multicolumn{2}{c}{Truck} & \multicolumn{2}{c}{Pedestrian} \ \\
    \cmidrule(r){2-3} \cmidrule(r){4-5} \cmidrule(r){6-7} \cmidrule(r){8-9}
    {Model} & {lidar} & {Camera}  & {0-50m} & {50-80m} & {0-50m}  & {50-80m} & {0-50m} & {50-80m} \\
    \midrule
  CP-VoxelNet$^{*}$ ~\cite{yin2020center} & \cmark &  & \textbf{95.3} & 28.1 & \textbf{71.7} & 12.3 & 94.1 & 17.3 \\ \hline
CP-PointPillars ~\cite{yin2020center}  & \cmark &  & 93.7 & 14.3 & 63.4 & 2.7 & 87.4 & 6.7 \\ \hline
PointPillars-FPN ~\cite{radosavovic2020designing} & \cmark & & 91.2 & 7.7 & 54.3 & 0.7 & 85.3 & 1.9 \\ \hline
PointPillars-SECFPN  ~\cite{radosavovic2020designing} & \cmark & & 89.7 & 5.8 & 62.4 & 1.4 & 81.3 & 1.9 \\ \hline
SSN-SECFPN ~\cite{zhu2020ssn} & \cmark &  & 91.1 & 8.3 & 61.4 & 1.2 & 78.1 & 1.0 \\ 
\Xhline{2\arrayrulewidth}
FCOS3D ~\cite{wang2021fcos3d} &  & \cmark & 87.2 & 37.0 & 48.5 & 8.1 & 67.9 & 12.6 \\ 
\Xhline{2\arrayrulewidth}
Bayesian Fusion ~\cite{chen2021multimodal} & \cmark & \cmark & 94.8 & 50.8 & 68.7 & 18.2 & 94.3 & 22.5 \\ \hline
CLOCs3D$^{**}$ & \cmark & \cmark &  94.8  & 50.8 & 68.7 & 18.2 & \textbf{94.3} & 22.5 \\ \hline
NMS Fusion (Ours) & \cmark & \cmark & 95.3 & 50.5 & 71.7 & 20.3 & 94.1 & \textbf{23.8} \\ \hline
AdaNMS Fusion (Ours) & \cmark & \cmark & 95.3 & 53.1 & 71.7 & 21.0 & 94.1 & 19.7 \\ \hline
MVP~\cite{yin2021multimodal} & \cmark & \cmark & \textbf{96.47} & 69.33 & \textbf{76.12} & 43.21 &  \textbf{96.08} & \textbf{58.89} \\ \hline
AdaNMS (MVP, FCOS3D) & \cmark & \cmark & \textbf{96.47} & \textbf{69.89} & \textbf{76.12} & \textbf{43.44} & \textbf{96.08} & 56.61 \\ \hline
\end{tabular}
\label{tab:verified_nuScenes_4m}
\end{table*}
}

\section{Zero lidar Point Objects}

\begin{figure}[t]
    \centering
        \includegraphics[width=.48\linewidth]{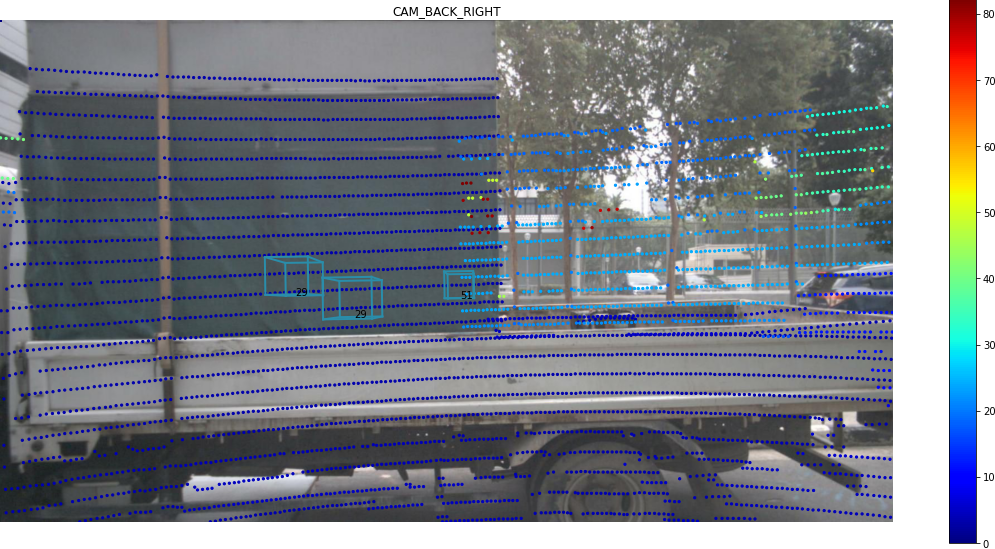}
        \includegraphics[width=.48\linewidth]{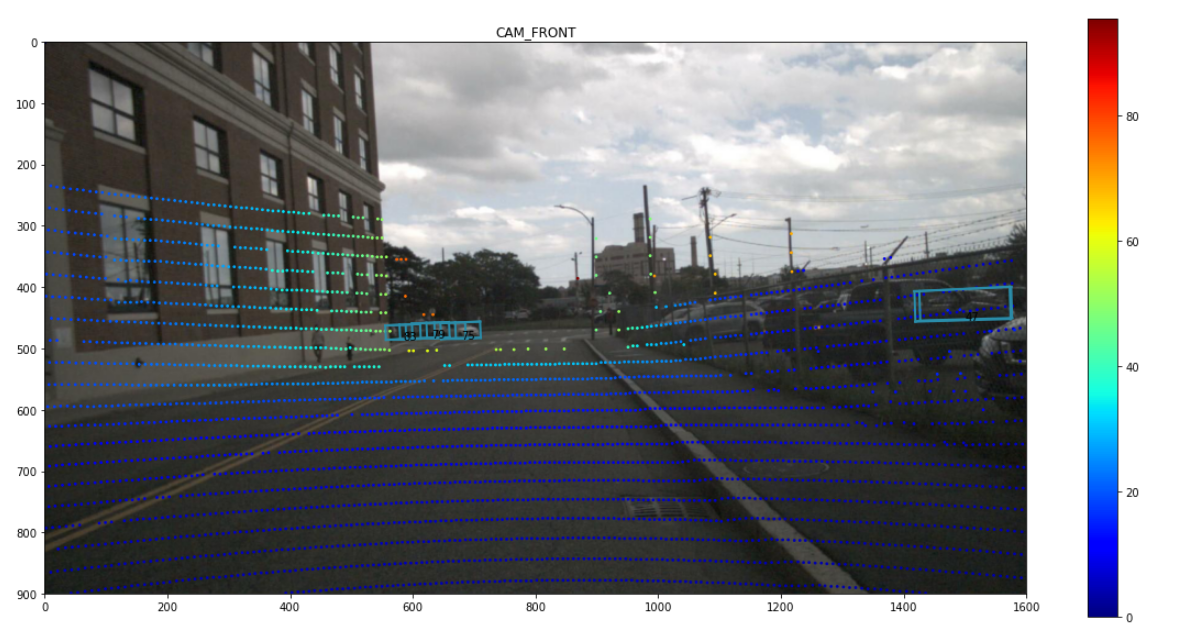}
    \vspace{-1mm}
    \caption{\small 
    Visualization of objects (annotated with bounding boxes) that have zero lidar points. Left: objects do not have lidar returns because they are occluded by a vehicle in front; they are not included in Far nuScenes validation set. Right: objects in the far side of road are unoccluded and are included in our Far nuScenes validation set.
    }
\vspace{-2mm}
\label{fig:zero-lidar1}
\end{figure}

\begin{figure}[t]
    \centering
        \includegraphics[width=.48\linewidth]{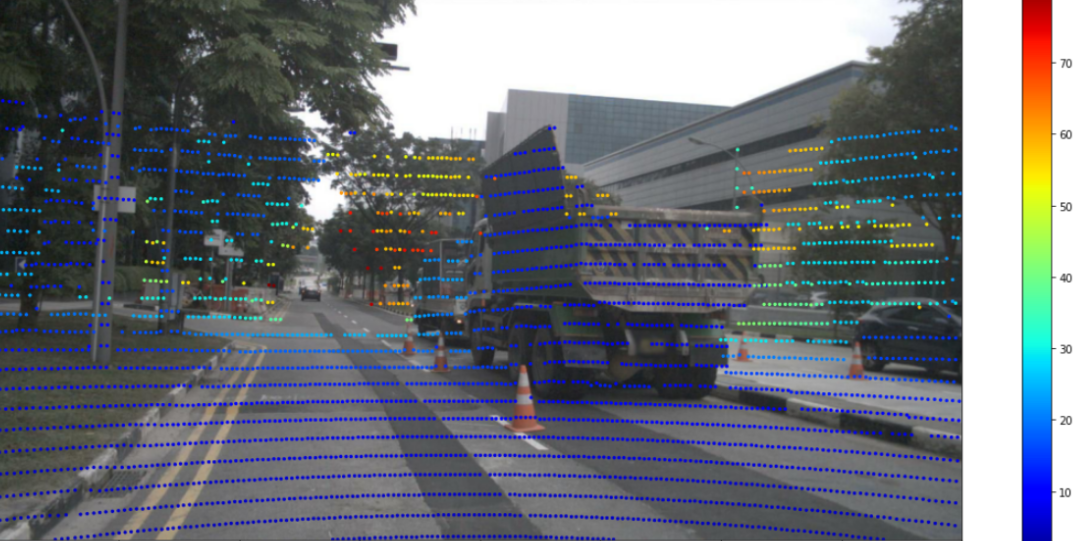}
        \includegraphics[width=.48\linewidth]{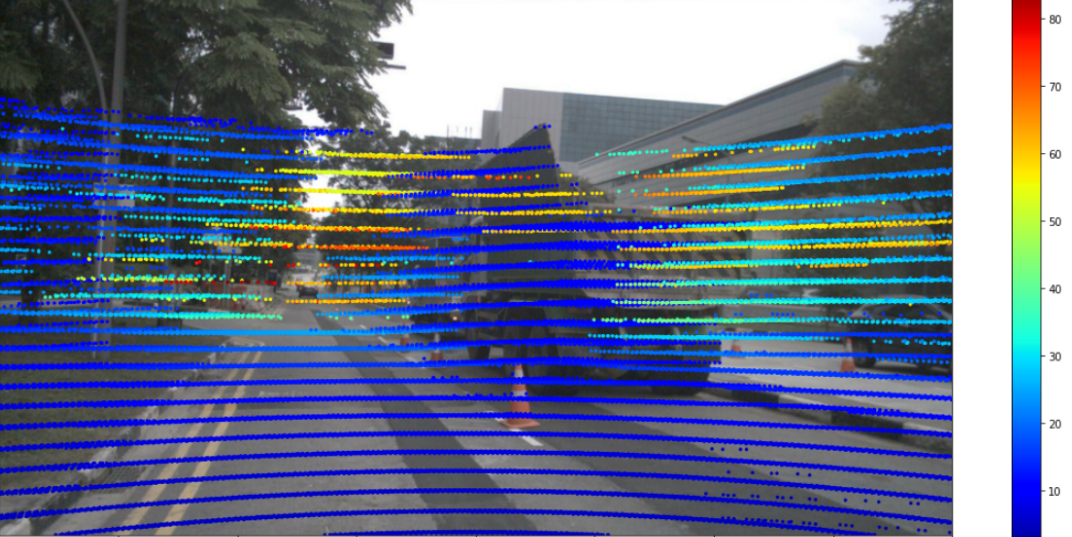}
    \vspace{-1mm}
    \caption{\small 
    Comparison between single lidar sweep and multiple lidar sweeps. The left image has objects with zero lidar points on the far side of the road whereas when we consider multiple sweeps (right image), the objects has non-zero number of points on it.
    }
\vspace{-4mm}
\label{fig:zero-lidar2}
\end{figure}

As discussed in Section 3.2 of the main paper, 
we include the unoccluded zero-lidar point objects in our validation set. Table~\ref{tab:anno_count_unoccluded} summarizes our findings for this data.
Fig.~\ref{fig:zero-lidar1} and \ref{fig:zero-lidar2} visualize some examples that have zero lidar points on far-field objects.
Table~\ref{tab:occluded_objects_suppl}  supplements Table 7 of main paper, comparing different methods on far-field objects that have zero lidar points.

{\setlength{\tabcolsep}{0.1em}
\begin{table*}[t]
\small
\centering
\caption{Annotations count for Zero-lidar point boxes in the nuScenes~\cite{Caesar_2020_CVPR} dataset. We can observe that the unoccluded lidar point objects account for $\geq$10\% of the annotations in the far-field (50-80m)}
\vspace{-2mm}
\begin{tabular}{c c c c}
\toprule
     & \multicolumn{3}{c}{50-80m} \\
     \cmidrule(r){2-4} 
\textbf{Dataset} & Non-zero lidar Point Boxes & Zero-lidar Point Boxes & Unoccluded Zero-lidar Point Boxes \\ 
\cmidrule(r){1-1}  \cmidrule(r){2-2} \cmidrule(r){3-3} \cmidrule(r){4-4}  
\textbf{Training} & 100123 & 45050 & 10430 \\ 
\textbf{Validation} & 21658 & 10069 & 2655 \\ 
\textbf{Far nuScenes (Validation)} & 10470 & 5126 & 1623 \\ 
\bottomrule
\end{tabular}
\label{tab:anno_count_unoccluded}
\end{table*}
}

{
\setlength{\tabcolsep}{0.1em} 
\begin{table}[t]
\scriptsize
\centering
\caption{\small
(This table supplements Table 7 of main paper.)
We include the unoccluded objects with zero lidar points using strategy described in 
Section 3.2
and calculate the mAP for 50-80m distance range. Clearly, our AdaNMS (MVP+FCOS3D) outperforms others on all categories except the Pedestrian and Trailer category. We see a slight decline in performance on this class as a distance adaptive IOU hurts the recall in a cluttered scene. Note, we don't include the classes Motorcycle, Bicycle and Traffic Cone as the number of annotations are quite low Table~\ref{tab:anno_count}.
}
\vspace{-2mm}
\begin{tabular}{c ccccccc} 
    \toprule
     & \multicolumn{7}{c}{50-80m} \\
    \cmidrule(r){2-8} 
    {Method} & {Car} & {Truck} & {Bus} & {Trailer} & {Construction}  & {Pedestrian} & {Barrier}\\
    &  &   &  &   & {Vehicle}  &  &  \\
    \midrule
        CP~\cite{yin2020center} & 20.5 & 8.6 & 0.1 & 1.1 & 0.0 & 11.2 & 18.0 \\ 
        FCOS3D~\cite{wang2021fcos3d} & 35.2 & 9.7 & 6.9 & 11.2 & 0.0 & 11.7 & 38.6  \\ 
        AdaNMS (CP, FCOS3D) & 45.2 & 18.5 & 7.1 & 5.4 & 0.0 & 14.4 & 41.4  \\ 
        \midrule
        MVP~\cite{yin2021multimodal} & 55.2 & 35.5 & 26.0 & \textbf{17.5} & \textbf{1.6} & \textbf{44.2} & 37.0 \\ 
        AdaNMS (MVP, FCOS3D) & \textbf{60.3} & \textbf{36.0} & \textbf{27.0} & 16.1 & \textbf{1.6} & 42.6 & \textbf{48.8} \\  
    
    \bottomrule
\end{tabular}
\label{tab:occluded_objects_suppl}
\end{table}
}

\begin{figure*}[t]
    \centering
    \includegraphics[width=1\linewidth, clip, trim={0cm 0.5cm 0cm 0.8cm}]{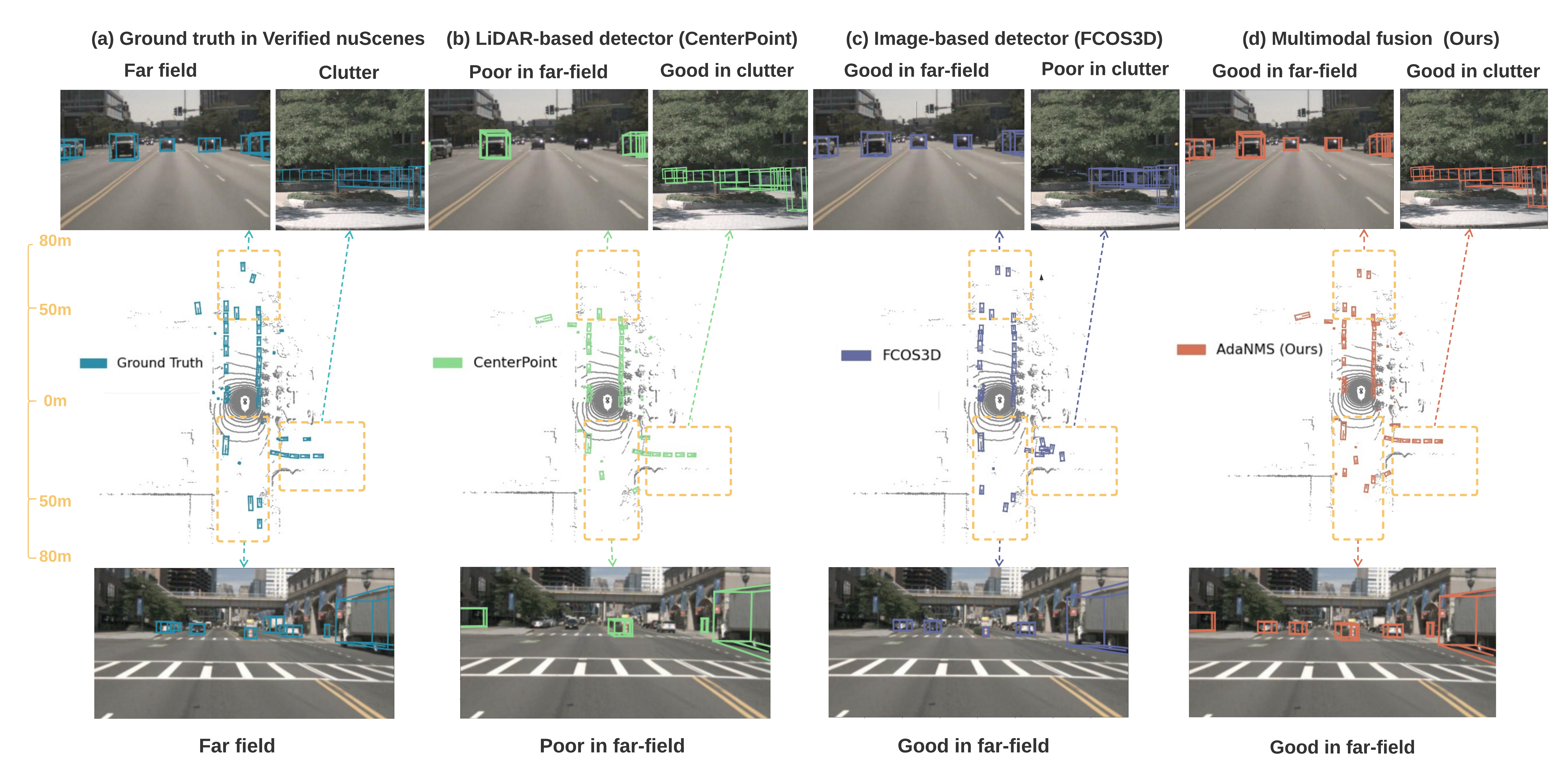}
    \vspace{-2mm}
    \caption{\small 
    Qualitative results on Far nuScenes. First, we note that our Far nuScenes contain clean annotations at a distance and for scenes with cluttered objects (a). By comparing the predictions between CenterPoint (b) and FCOS3D (c), we observe the image-based FCOS3D contains higher quality predictions for Far3Det, 
    which is not reflected in the standard evaluation (Table 3 in the main paper).
    Our proposed fusion method AdaNMS (d) leverages their respective advantages and greatly improves detection of far-field objects.
    }
\label{fig:qualitative}
\vspace{-2mm}
\end{figure*}

\section{Visualization, Demo, and Code}

Fig.~\ref{fig:qualitative} supplements the Fig. 4 in the main paper, demonstrating  a ``failure mode'' of image-based detectors as below. 
The detections are quite good in the projected image space but might have notable errors in the BEV. Existing evaluation protocols penalize such errors unreasonably too heavily, hence motivating our metrics that use distance adaptive thresholds.

We also attach a video demo ({\tt demo.mp4}) in this supplemental material, visualizing how our method improve 3D detections in the far-field. We further attach our code. We will publish them along our paper.

\end{document}